\documentclass[letterpaper]{article} 
\usepackage{aaai2027}  
\usepackage[hyphens]{url}  
\usepackage{graphicx} 
\usepackage{natbib}  
\usepackage{caption} 

\usepackage{afterpage}
\usepackage{cuted}
\usepackage{aaai2027-cuted-compat}
\usepackage{booktabs}
\usepackage{amsmath}
\usepackage{amssymb}
\usepackage{xcolor}
\definecolor{captionorange}{RGB}{230,126,34}
\title{UniMoFlow: Grounding Instruction-Driven 3D Human Motion Editing in Generation}
\author{
Yilei Hua\textsuperscript{\rm 1},
Beibei Jing\textsuperscript{\rm 2},
Ce Zheng\textsuperscript{\rm 1},
Hanyu Zhou\textsuperscript{\rm 3,4},
Yawei Luo\textsuperscript{\rm 5},
Wei Yang\textsuperscript{\rm 1,*}
}
\affiliations{
\textsuperscript{\rm 1}School of Computer Science, Huazhong University of Science and Technology\\
\textsuperscript{\rm 2}Nanyang Normal University\qquad 
\textsuperscript{\rm 3}School of Computing, National University of Singapore\\
\textsuperscript{\rm 4}Forma AI, Xingtai Qiyuan Co., Ltd.\qquad 
\textsuperscript{\rm 5}School of Software Technology, Zhejiang University\\
\textsuperscript{\rm 2}20261123@nynu.edu.cn\qquad \textsuperscript{*}Corresponding author
}

\begin{document}

\maketitle
\vspace{0pt}

\begingroup\setlength{\stripsep}{0pt}\begin{strip}
    \centering
    \includegraphics[width=0.95\textwidth]{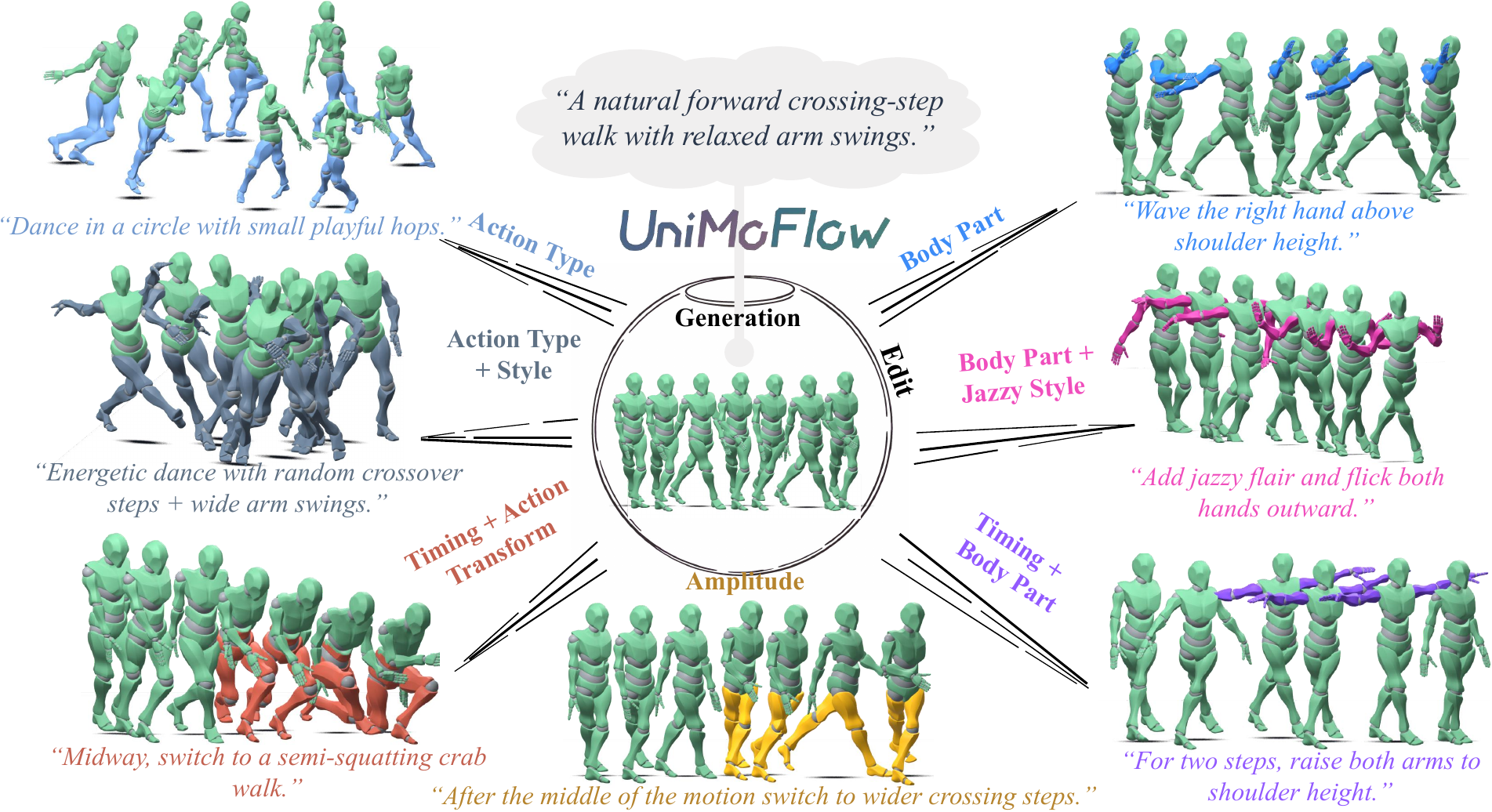}
    \vspace{0.25em}
    \captionof{figure}{We propose \textbf{UniMoFlow}, a unified method for high-quality motion generation and editing. Given a generated or user-provided source motion sequence, UniMoFlow adjusts body-part movements, action types, timing, amplitude, style, and compound edits while faithfully following textual instructions.}
    \label{fig:teaser}\vspace*{4em}
    \vspace{0pt}
\end{strip}\endgroup\relax

\begin{abstract}
Instruction-driven editing of 3D human motion requires precise spatiotemporal localization, rich semantic grounding, and strict preservation of unmodified content. Existing methods either resort to training-free adaptation of generative models or rely solely on triplet supervision; however, adaptation often yields suboptimal control, and manually curated triplet datasets remain severely limited in scale and semantic diversity. To overcome this bottleneck, we ground motion editing directly within text-to-motion generation across data, architecture, and inference. At the data level, we develop a closed-loop synthesis-and-verification pipeline that produces \textbf{Omni-MoEdit}, a large-scale dataset spanning body-part, amplitude, temporal, action, and style edits. At the architectural level, we introduce \textbf{UniMoFlow}, a unified latent flow-matching model that shares broad semantic and kinematic knowledge between generation and editing. At the inference level, \textbf{SAFE} (Source-Anchored Flow Editing) complements UniMoFlow with controllable, source-anchored refinement. Furthermore, we augment standard evaluations with semantics-aware metrics to account for valid edits that inherently deviate from a single ground-truth reference. Extensive experiments demonstrate improved target-text alignment, edit effectiveness, and cycle consistency, while maintaining competitive source fidelity and text-to-motion generation quality.

\end{abstract}

\begin{center}
\textbf{Code:} \url{https://github.com/Yilei-Hua/UniMoFlow}
\end{center}
\vspace{-0.5em}

\section{Introduction}
Human motion generation and editing are central to animation, virtual reality, and embodied agents. Although recent text-to-motion models synthesize plausible motion from natural language \citep{guo2022generating,tevet2023human,chen2023executing,zhang2023generating,guo2024momask}, a single pass seldom meets precise requirements for timing, articulation, intensity, or style. Practical creation therefore requires iterative natural-language editing that localizes the requested change, grounds its semantics, and preserves irrelevant source structure, as illustrated in Figure~\ref{fig:teaser}.

Instruction-driven editing requires strong generative priors: interpreting \textit{``perform the jump in an exhausted manner''} requires linking an abstract style to its kinematic realization. Current methods nevertheless face two structural limitations. First, they commonly treat editing as an auxiliary task decoupled from text-to-motion generation rather than as a native capability of one generator. Supervised editors \citep{athanasiou2024motionfix,jiang2025motionrefit,li2025simmotionedit,yang2025partmotionedit,guo2025motionlab} optimize source--instruction--target triplet mappings but do not jointly absorb broad text-to-motion supervision, whereas training-free methods \citep{kulikov2025flowedit,chen2024motionclr,hong2025salad} manipulate pretrained diffusion or flow priors only at inference without learning an explicit instruction-to-modification mapping. Second, mainstream datasets retrieve structurally similar clips and annotate their differences, as in MotionFix \citep{athanasiou2024motionfix}. This retrieval constraint favors short, localized pose or amplitude adjustments: pairs with large differences are hard to align, while nearly identical pairs offer little editing signal. It therefore limits both dataset scale and coverage of action, temporal, and stylistic edits. 

To bridge this gap, we ground motion editing in text-to-motion generation across data, architecture, and inference.
At the data level, we distill expressive text-to-motion data into scalable editing supervision. Qwen3-8B \citep{yang2025qwen3} converts SnapMoGen source captions \citep{guo2025snapmogen} into edit instructions and target descriptions spanning body-part, amplitude, temporal, action, and style changes. A base DiT trained only for text-to-motion then uses classical FlowEdit \citep{kulikov2025flowedit} to synthesize target motions. An evaluator retains pairs with high target-motion--target-text alignment and a clear improvement over the source. This pipeline yields \textbf{Omni-MoEdit}, a larger and more diverse edit dataset whose multi-stage synthesis can be amortized by the final editor in one conditional forward pass \citep{brooks2023instructpix2pix}.
Architecturally, we introduce \textbf{UniMoFlow}, a unified latent flow-matching framework for text-to-motion generation and instruction-driven editing. It places captions or edit instructions, source motion, and denoising states in one self-attention context, with lightweight task-specific heads for the two modes. Joint training lets high-quality generation data regularize imperfect synthetic edit pairs, while editing supervision strengthens language-motion grounding; full token-level source access supports precise spatiotemporal edits.
At inference, \textbf{SAFE (Source-Anchored Flow Editing)} complements rather than replaces UniMoFlow's native editor and is not part of data synthesis. It reuses the learned flow field to refine a source motion while anchoring its trajectory and steering it with the instruction. Its continuous strength parameter controls the edit--preservation trade-off without retraining, particularly for fine-grained adjustments.
Single synthetic targets incompletely evaluate editing: valid edits may differ from the reference, whereas nearly unchanged motions can score well on reconstruction metrics. We therefore combine GT-reference measures with target-text-driven criteria. Besides R-Precision, Matching Score, and FID, we track improvement over the source with respect to the target description, source preservation, edit localization, positive-improvement ratio, and reverse-edit cycle consistency. This separates semantic success, source fidelity, and actual edit effect.

In summary, our contributions are threefold:
\begin{itemize}
\setlength{\itemsep}{0.15em}
\setlength{\parsep}{0pt}
\setlength{\topsep}{0.25em}
\item We develop a generation-driven synthesis-and-filtering pipeline to construct Omni-MoEdit, a dataset that distills the expressive text-motion diversity of SnapMoGen into scalable editing supervision spanning body-part, amplitude, temporal, action, and style edits.
\item We propose UniMoFlow, a unified latent flow-matching framework that intrinsically grounds instruction-driven editing in text-to-motion generation. This is complemented by SAFE, an inversion-free UniMoFlow sampling mode providing continuous control over editing strength.
\item We introduce semantics-aware evaluation metrics to better capture the multi-modal nature of motion editing, and demonstrate that UniMoFlow substantially outperforms prior editors, while maintaining highly competitive text-to-motion generation quality.
\end{itemize}

\section{Related Work}

\noindent \textbf{Text-to-Motion Generation}
Text-to-motion generation synthesizes 3D human motion from free-form language and is commonly divided into continuous-space and discrete-token routes. Continuous approaches directly generate joint trajectories or continuous VAE latents, ranging from VAE-based TEMOS to diffusion and latent-diffusion models such as MotionDiffuse, MDM, and MLD \citep{guo2022generating,petrovich2022temos,zhang2024motiondiffuse,tevet2023human,chen2023executing}. Gaussian diffusion and flow matching both define generative processes over continuous variables; UniMoFlow follows this route and learns a flow field in continuous latent motion space.
Discrete approaches first quantize motion into learned codes, then perform sequence generation or masked-token reconstruction: TM2T, T2M-GPT, and MotionGPT model token sequences, whereas MoMask and MMM recover masked motion codes \citep{guo2022tm2t,zhang2023generating,jiang2023motiongpt,guo2024momask,pinyoanuntapong2024mmm}. In contrast, UniMoFlow retains continuous latent states, which permits a common flow-matching formulation for both text-conditioned generation and source-conditioned editing.

\noindent \textbf{Instruction-Driven Motion Editing}
Instruction-driven motion editing modifies a source motion according to a natural-language command while preserving irrelevant structure. MotionFix established this setting with source--target--instruction triplets and TMED \citep{athanasiou2024motionfix}. Follow-up supervised editors mainly refine source conditioning: MotionReFit and SimMotionEdit inject compressed source-motion cues, PartMotionEdit targets body-part control, OmniME introduces positive--negative supervision, and MotionLab explores flow-based editing \citep{jiang2025motionrefit,li2025simmotionedit,yang2025partmotionedit,shi2026omnime,guo2025motionlab}. These architectural studies remain largely trained on MotionFix-style pairs, whose limited scale and localized edit distribution leave broader edit-pair synthesis underexplored.
A complementary line transfers pretrained priors or avoids edit training altogether. OmniMoGen and UMO fine-tune large pretrained language or motion models on MotionFix-style supervision, while MotionMaster enables zero-shot editing through large-scale motion--text modeling with discrete motion tokens \citep{bu2025omnimogen,cong2026umo,jiang2026motionmaster}. Training-free editors such as FlowEdit, MotionCLR, and SALAD manipulate pretrained generators at inference \citep{kulikov2025flowedit,chen2024motionclr,hong2025salad}. For fair comparison, our current UniMoFlow instantiation does not rely on additional priors from open-source foundation models. For zero-shot or training-free editing, we mainly use such samplers to synthesize candidate pairs with Qwen3-8B, thereby supporting the training of native editing ability while preserving continuous-space generation.

\noindent \textbf{Context-Aware Editing and Generation}
A related trend in visual generation trains native instruction editors from scalable synthetic or weakly curated edit pairs. InstructPix2Pix combines LLM-generated instructions with synthetic image pairs \citep{brooks2023instructpix2pix}; more recent systems, including FLUX.1 Kontext, DreamOmni, VIBE, ACE, OmniGen2, Step1X-Edit, and HiDream-E1, further demonstrate that data pipelines and context-aware architectures can jointly support generation and editing \citep{blackforestlabs2025fluxkontext,xia2024dreamomni,alekseenko2026vibe,han2024ace,wu2025omnigen2,liu2025step1xedit,cai2025hidream}.
We transfer this principle to 3D human motion: Qwen3-8B produces edit instructions and target descriptions, a text-to-motion DiT with FlowEdit synthesizes target motions, and automatic filtering selects usable pairs. Rather than relying on synthetic pairs alone, UniMoFlow jointly trains on high-quality generation data, which regularizes residual synthesis artifacts while the edit triplets strengthen language--motion grounding.

\begin{figure}[t]
    \centering
    \includegraphics[width=\columnwidth]{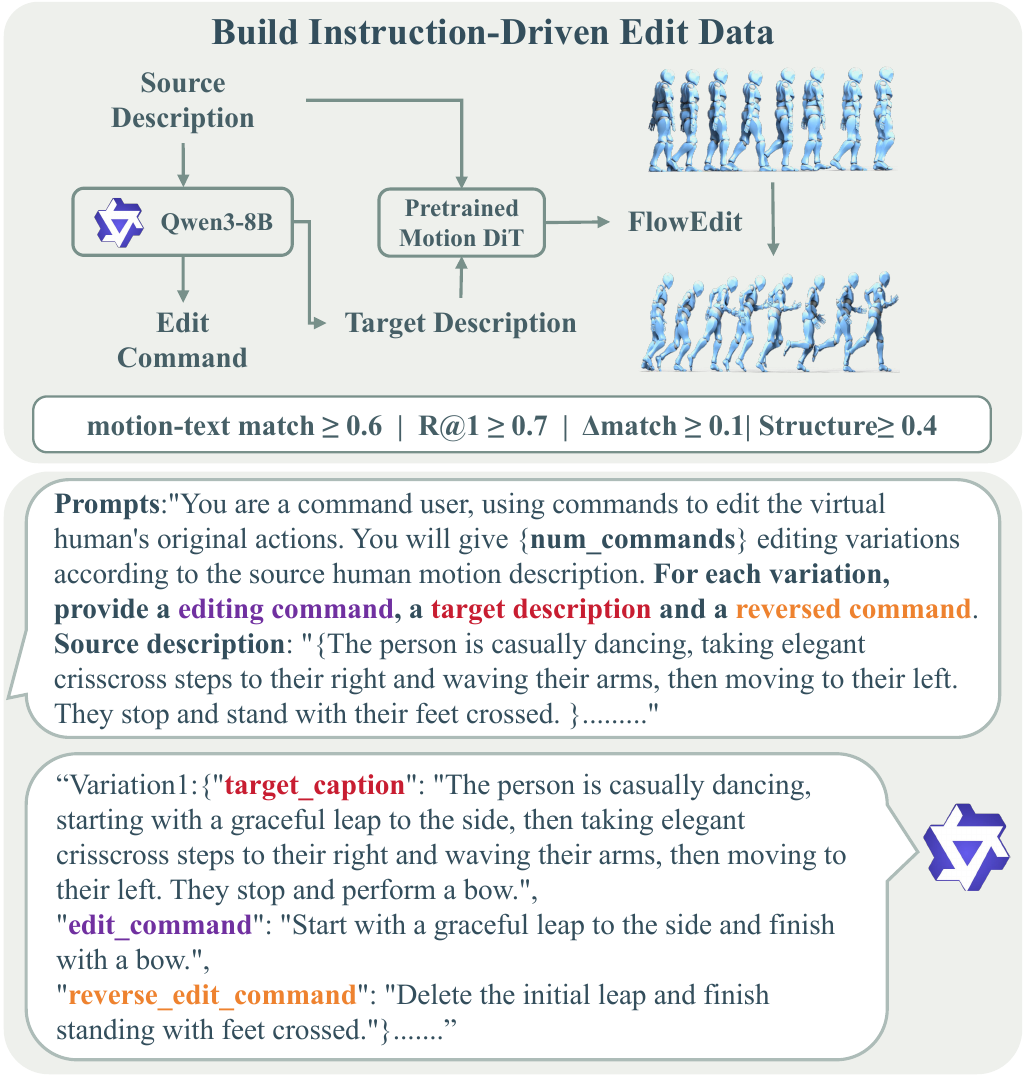}
    \vspace{-4pt}
    \caption{The Omni-MoEdit data synthesis pipeline. The upper part gives a simplified workflow of edit-data synthesis, while the lower part shows how pretrained Qwen is prompted to generate task-specific edits.}

    \label{fig:data-synthesis-pipeline}
\end{figure}

\section{The Omni-MoEdit Dataset}
\begin{table}[t]
\centering
\small
\begin{tabular}{l r c c c}
\toprule
Dataset & Pairs & Src. text & Tgt. text & Rev. inst. \\
\midrule
MotionFix & 6,730 & No & No & No \\
STANCE-Adj. & 16,000 & No & No & No \\
Omni-MoEdit & 55,641 & Yes & Yes & Yes \\
\bottomrule
\end{tabular}
\vspace{2pt}
\caption{Comparison with motion editing datasets.}
\label{tab:dataset-comparison}
\end{table}

Supervised motion editing relies on \textsc{<source motion, edit instruction, target motion>} triplets, yet existing datasets, constructed by retrieving similar motion-capture clip pairs, are limited in scale and biased toward local pose and amplitude adjustments. 
A valid editing triplet must satisfy two constraints: the target motion must realize the instructed modification (\emph{edit fidelity}) while remaining consistent with the source in all unedited respects (\emph{source preservation}). Naturally occurring pairs satisfying both are rare, we synthesize triplets and enforce both constraints through automatic filtering. We build on SnapMoGen \citep{guo2025snapmogen} because its fine-grained captions permit deriving precise instructions as deltas between descriptions, its motions span diverse actions and styles, and its pretrained text--motion alignment evaluator provides the required filter.

\begin{figure*}[t]
    \centering
    \includegraphics[width=\textwidth]{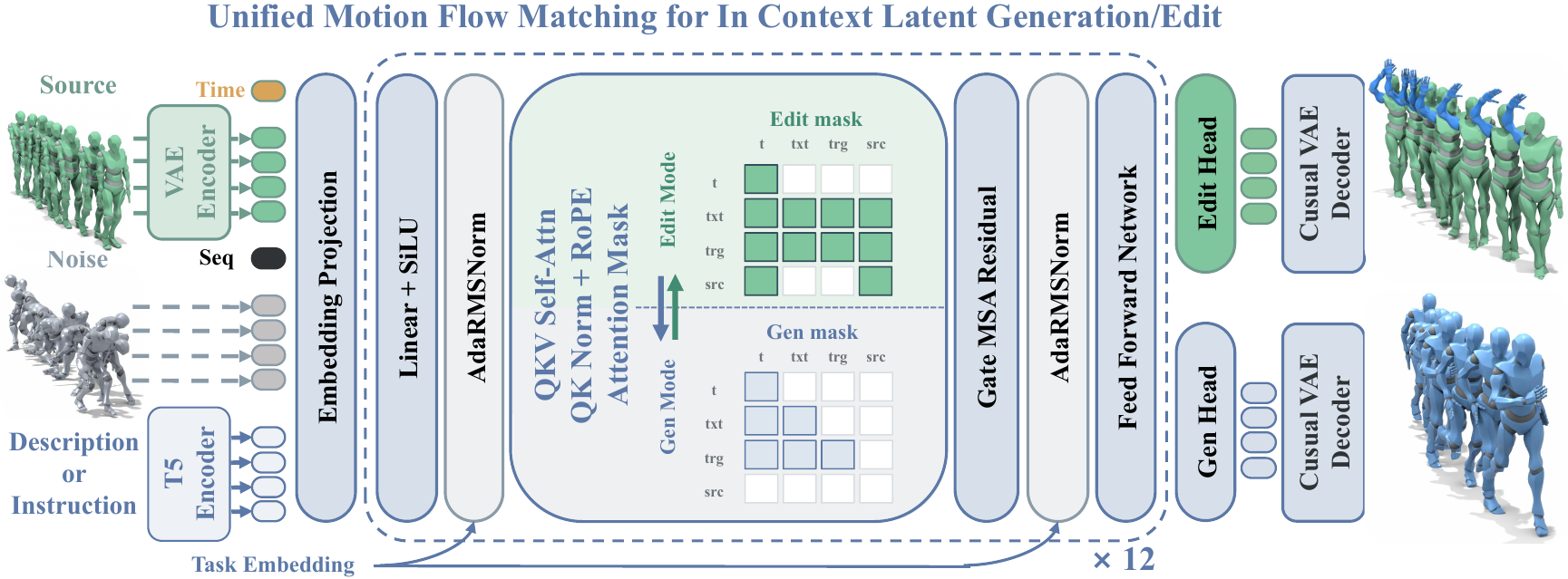}
    \vspace{-4pt}
        \caption{Architecture of UniMoFlow. Text, noisy target-motion, and optional full source-motion tokens form a unified context sequence. Mode-specific global self-attention masks switch the shared flow-matching Transformer between generation and native editing, while lightweight task heads predict the target velocity.}
    \label{fig:UniMoFlow-framework}
    \vspace{-1em}
\end{figure*}

 
The pipeline comprises three stages, illustrated in Figure~\ref{fig:data-synthesis-pipeline}.
 
\noindent \textbf{(i) Synthesis backbone.}
Following SnapMoGen, we train a causal VAE tokenizer, adapted from the Wan-VAE video architecture \citep{wang2025wan}, that compresses every four motion frames into one continuous latent token. A DiT-style flow-matching text-to-motion model \citep{peebles2023dit}, trained only for text-conditioned motion generation in this latent space, serves as the target-motion synthesis backbone.
 
\noindent \textbf{(ii) Edit candidate generation.}
For each source caption, Qwen3-8B \citep{yang2025qwen3} produces $N$ candidate pairs, each comprising a forward edit instruction and a target description. The prompt constrains each candidate to a single edit drawn from a taxonomy covering edit timing, body parts, modification type, and style, and asks the target description to stay close to the source description whenever possible, while permitting a focused, semantically coherent instruction to specify local, action-level, stylistic, or coupled changes. Qwen3-8B additionally produces the corresponding reverse instruction. The prompt schema and split protocol are detailed in Supplementary \emph{Omni-MoEdit Dataset Details}.
 
\noindent \textbf{(iii) Target-motion synthesis and filtering.}
For each candidate, the base text-to-motion DiT synthesizes a target motion using classical FlowEdit sampling \citep{kulikov2025flowedit}, conditioned on the source motion and the Qwen-generated target description. SAFE is used only as a complementary inference mode for UniMoFlow and is not involved in data synthesis. We retain a candidate only if (a) the Matching Score between its synthesized target motion and target description exceeds 0.6, (b) target-text retrieval satisfies $\mathrm{R}@1 \ge 0.7$, (c) the target Matching Score is at least $\tau=0.1$ higher than the Matching Score between the source motion and target description, and (d) its structure score is at least 0.4. These constraints jointly enforce target-text alignment, a measurable semantic change from the source, and basic motion structure. Finally, Omni-MoEdit contains 46{,}911 / 2{,}895 / 5{,}835 edit triplets in the train / validation / test splits, built from 22{,}941 unique source motions and 44{,}836 synthesized target motions.   Table~\ref{tab:dataset-comparison} compares Omni-MoEdit with MotionFix and STANCE Adjustment from MotionReFit \citep{jiang2025motionrefit}. STANCE Adjustment contains 16,000 fine-grained adjustment triplets, while Omni-MoEdit contains 55,641 pairs, about 3.5$\times$ larger, and further provides source captions, target captions, and reverse instructions. Omni-MoEdit supports fine-grained edits as well as coarser action-type and style-transfer edits.

\section{Unified Motion Generation and Editing}

The central design of UniMoFlow is to learn from synthesized edit triplets without isolating editing from the stronger text-to-motion prior. The synthetic triplets supervise instruction-conditioned source-to-target modifications, while high-quality text-to-motion data supplies cleaner semantic grounding and regularizes residual artifacts in automatically synthesized edits. Conversely, instruction-driven triplets sharpen the correspondence between language and motion, benefiting generation as well. Rather than attaching a separate editing branch or compressing the source motion into a single condition vector, UniMoFlow assembles the text condition, noisy target motion, and the full source-motion token sequence into a shared context. \textbf{A mode-specific global self-attention mask then switches the same backbone between generation and editing.} The shared Transformer therefore learns language--motion grounding, source conditioning, and flow dynamics across both tasks, while lightweight task-specific heads specialize only the final velocity prediction.

\subsection{Unified Token Sequence}\label{sec:context}
Let $\mathbf{x}_s$ and $\mathbf{x}_t$ denote the source and target motion sequences, $c$ the text condition, and $\tau$ the continuous flow time step. We encode both motions with the pretrained causal VAE used for dataset construction, yielding latent sequences $\mathbf{z}_s$ and $\mathbf{z}_t$. At time $\tau$, the target latent is noised as $\mathbf{z}_t^\tau=(1-\tau)\mathbf{z}_t+\tau\boldsymbol{\epsilon}$. A frozen pretrained T5 encoder maps $c$ to text tokens $\mathbf{Z}_c$, while the time step is mapped by a sinusoidal embedding and an MLP to a time embedding $\mathbf{z}_\tau$.

All modalities are projected into the shared hidden space before sequence assembly. Crucially, modality tags are not concatenated as additional tokens. Instead, a time-conditioned tag is added to every token of its corresponding modality: $\widehat{\mathbf{Z}}_c^{m}=\mathbf{Z}_c+\mathcal{T}_c^{m}(\mathbf{z}_\tau)$ for text tokens with $m\in\{\mathrm{edit},\mathrm{gen}\}$, $\widehat{\mathbf{Z}}_t^{\tau}=\mathbf{Z}_t^{\tau}+\mathcal{T}_t(\mathbf{z}_\tau)$ for target tokens, and $\widehat{\mathbf{Z}}_s=\mathbf{Z}_s+\mathcal{T}_s(\mathbf{z}_\tau)$ for source tokens, where each tag is broadcast over its token length. Here, $\mathcal{T}_c^{m}$ is a mode-specific text-tag projection, and $\mathcal{T}_t$ and $\mathcal{T}_s$ are target- and source-tag projections. Thus, text, target, and source tokens carry their respective type tags by addition rather than by concatenation.

In editing mode, the resulting context sequence is
\begin{equation}
\mathbf{Z}_{\mathrm{edit}}=\big[\mathbf{z}_\tau,\widehat{\mathbf{Z}}_c^{\mathrm{edit}},\widehat{\mathbf{Z}}_t^{\tau},\mathbf{z}_{\mathrm{sep}},\widehat{\mathbf{Z}}_s\big],
\label{eq:seq_edit}
\end{equation}
where $\mathbf{z}_{\mathrm{sep}}$ is a learnable separator. In generation mode, the source branch and separator are omitted:
\begin{equation}
\mathbf{Z}_{\mathrm{gen}}=\big[\mathbf{z}_\tau,\widehat{\mathbf{Z}}_c^{\mathrm{gen}},\widehat{\mathbf{Z}}_t^{\tau}\big].
\label{eq:seq_gen}
\end{equation}
Finally, a learnable task embedding $\mathbf{e}_m$ is added to $\mathbf{z}_\tau$ to form $\mathbf{c}_m=\mathbf{z}_\tau+\mathbf{e}_m$, which modulates the Transformer blocks through adaptive layer normalization. This construction retains a shared token interface for generation and editing while distinguishing their roles through additive tags and the mode-specific attention mask.

\begin{table*}[!t]
\centering
\scriptsize
\renewcommand{\arraystretch}{0.78}
\setlength{\tabcolsep}{3pt}
\resizebox{\textwidth}{!}{
\begin{tabular}{lccccccc}
\toprule
Method & TR@1 $\uparrow$ & TR@2 $\uparrow$ & TR@3 $\uparrow$ & Match $\uparrow$ & Struct $\uparrow$ & Region $\uparrow$ & PosRatio $\uparrow$ \\
\midrule
GT & 0.8973 & 0.9806 & 0.9953 & 0.7696 & -- & -- & 100.0\% \\
UniMoFlow (SAFE) & \textbf{0.6347} & \textbf{0.7972} & \textbf{0.8651} & \textbf{0.5861} & \underline{0.7927} & \underline{0.3384} & \textbf{72.8\%} \\
UniMoFlow (Native) & \underline{0.5492} & \underline{0.7036} & \underline{0.7777} & \underline{0.5263} & 0.3725 & \textbf{0.3886} & \underline{54.2\%} \\
OmniME (CVPR'26) & 0.3751 & 0.5502 & 0.6479 & 0.3855 & 0.5595 & 0.2355 & 24.3\% \\
MotionLab (ICCV'25) & 0.4839 & 0.6652 & 0.7593 & 0.5083 & 0.7000 & 0.2749 & 48.4\% \\
SimMotionEdit (CVPR'25) & 0.1290 & 0.2250 & 0.2932 & 0.1485 & 0.0760 & 0.2154 & 11.0\% \\
MotionReFit (CVPR'25) & 0.1556 & 0.2659 & 0.3470 & 0.1484 & 0.2570 & 0.1019 & 4.7\% \\
TMED (SIGGRAPH Asia'24) & 0.4262 & 0.6243 & 0.7362 & 0.4552 & \textbf{0.8759} & 0.1990 & 27.5\% \\
\bottomrule
\end{tabular}}
\caption{Editing performance on Omni-MoEdit using target-text-centered metrics. Struct and Region are not defined for GT. Bold and underline denote best and second-best results.}
\label{tab:edit-target-text}
\vspace{-0.8em}
\end{table*}

\begin{table}[!t]
\centering
\footnotesize
\renewcommand{\arraystretch}{0.96}
\setlength{\tabcolsep}{2.5pt}
\resizebox{\columnwidth}{!}{
\begin{tabular}{lcccc}
\toprule
Method & GT R@1 $\uparrow$ & Src. R@1 $\uparrow$ & Cycle-Con $\uparrow$ & FID $\downarrow$ \\
\midrule
UniMoFlow (S) & \textbf{0.6762} & \underline{0.5909} & \textbf{0.4458} & \textbf{12.45} \\
UniMoFlow (N) & \underline{0.5852} & 0.3664 & \underline{0.2283} & \underline{13.37} \\
OmniME & 0.3847 & 0.4807 & 0.0711 & 115.92 \\
MotionLab & 0.5197 & 0.5641 & 0.2190 & 26.06 \\
SimMotionEdit & 0.1208 & 0.0728 & 0.0050 & 173.72 \\
MotionReFit & 0.1168 & 0.1777 & 0.0097 & 692.13 \\
TMED & 0.4579 & \textbf{0.6166} & 0.1071 & 86.04 \\
\bottomrule
\end{tabular}}
\caption{Editing performance on Omni-MoEdit under the MotionFix-style ground-truth protocol. GT/Src. R@1 measure retrieval against the synthesized target/source, FID is computed against the synthesized target, and Cycle-Con is our reverse-edit consistency metric.}
\label{tab:edit-gt}
\vspace{0.35em}
\end{table}

\subsection{The UniMoFlow Model}\label{sec:objective}

UniMoFlow employs a single shared Transformer denoiser with mode-specific output heads. Each block applies global self-attention over the unified token sequence, rotary positional embeddings, and adaptive layer normalization modulated by the time and task embeddings. In contrast to the base DiT synthesis backbone \citep{peebles2023dit}, UniMoFlow contains no text cross-attention: the text condition, the noisy target latents, and, in editing mode, the source latents reside in a single sequence (Eqs.~\ref{eq:seq_edit}--\ref{eq:seq_gen}) and interact exclusively through self-attention, i.e., in-context conditioning. This design is not merely a simplification: compared with the cross-attention Base DiT baseline, UniMoFlow improves FID from $15.884$ to $15.331$ and Matching Score from $0.705$ to $0.716$ on SnapMoGen (Table~\ref{tab:t2m-snapmogen}), indicating that text injection through the shared self-attention context is more effective for generation. It also extends naturally to source-aware editing, since source tokens participate in the same attention operation as all other conditions rather than being compressed into a fixed memory. After the final block, UniMoFlow extracts the target-token slice of the sequence and predicts the flow velocity through the generation or editing head according to the task tag.

For a target latent motion $\mathbf{z}$ and Gaussian noise $\boldsymbol{\epsilon}$, we use the standard rectified-flow interpolation
\begin{equation}
\mathbf{z}^{\,t}=(1-t)\boldsymbol{\epsilon}+t\mathbf{z},\qquad
\frac{\mathrm{d}\mathbf{z}^{\,t}}{\mathrm{d}t}=\mathbf{z}-\boldsymbol{\epsilon}.
\label{eq:interp}
\end{equation}
Motion sequences vary substantially in length, and longer sequences retain more recoverable signal at a given noise level; a single time schedule is therefore suboptimal across lengths. Analogous to resolution-dependent timestep shifting in image synthesis \citep{esser2024scaling}, we condition the model on a length-dependent shifted time $\tau = s(t;\ell)$, where $\ell$ is the latent sequence length and $s(\cdot\,;\ell)$ is a monotone reparameterization with Jacobian $J(t;\ell) = \partial \tau / \partial t > 0$. Since the interpolation in Eq.~\ref{eq:interp} remains linear in $t$, the chain rule gives the supervised velocity in shifted time:
\begin{equation}
    {v}^{*}
    = \frac{\mathrm{d}\mathbf{z}^{\,t}}{\mathrm{d}\tau}
    = \frac{\mathrm{d}\mathbf{z}^{\,t}}{\mathrm{d}t} \cdot \frac{\partial t}{\partial \tau}
    = \frac{\mathbf{z} - \boldsymbol{\epsilon}}{J(t;\ell)}.
\label{eq:shifted_velocity}
\end{equation}
Let $v_\theta$ denote the UniMoFlow velocity predictor, taking as input the unified token sequence of the corresponding mode. Training alternates editing samples, drawn from Omni-MoEdit triplets, and generation samples, drawn from the text-to-motion corpus. The objective is
\begin{equation}
\begin{aligned}
\mathcal{L}
= \mathbb{E}
\big\|  {v}_\theta(\mathbf{Z}_\text{edit}) - {v}^{*}  \big\|_2^2  +
 \lambda \, \mathbb{E}  \big\| v_\theta(\mathbf{Z}_\text{gen})- {v}^{*}  \big\|_2^2,
\end{aligned}
\label{eq:objective}
\end{equation}
The text condition is replaced with a null embedding with probability $p_{\mathrm{drop}}$ in both modes; in editing mode, the source condition is retained under text dropout, which yields the source-anchored null direction exploited by SAFE at inference (Section~\ref{sec:anchorflow}).

\begin{table}[!t]
\centering
\footnotesize
\renewcommand{\arraystretch}{0.94}
\setlength{\tabcolsep}{3pt}
\begin{tabular}{@{}lcccc@{}}
\toprule
Method & R@3 $\uparrow$ & FID $\downarrow$ & Match $\uparrow$ & MM. $\uparrow$ \\
\midrule
Real motions & 0.985 & 0.001 & 0.837 & -- \\
\midrule
\multicolumn{5}{c}{Discrete Methods} \\
\midrule
T2M-GPT (CVPR'23) & 0.812 & 32.629 & 0.573 & \textbf{9.172} \\
MoMask (CVPR'24) & \underline{0.927} & \underline{17.404} & \underline{0.664} & \underline{8.183}\\
MoMask++ (NeurIPS'25) & \textbf{0.938} & \textbf{15.060} & \textbf{0.685} & 7.259\\
\midrule
\multicolumn{5}{c}{Continuous Methods} \\
\midrule
MDM (ICLR'23) & 0.727 & 57.783 & 0.481 & \textbf{13.412} \\
StableMoFusion (ACM MM'24) & 0.888 & 27.801 & 0.605 & 9.064 \\
MARDM (CVPR'25) & 0.860 & 26.878 & 0.602 & 9.812 \\
Unified Flow & \textbf{0.987} & 16.567 & 0.663 & \underline{11.259} \\
Base DiT & 0.922 & \underline{15.884} & \underline{0.705} & 6.569 \\
UniMoFlow (edit+gen) & \underline{0.930} & \textbf{15.331} & \textbf{0.716} & 6.651 \\
\bottomrule
\end{tabular}
\caption{Text-to-motion generation on the SnapMoGen test split. We report R-Precision Top-3, FID, Matching Score, and MModality. Bold and underline denote best and second-best results within each method group.}
\label{tab:t2m-snapmogen}
\vspace{0.75em}
\end{table}

\subsection{SAFE: Source-Anchored Flow Editing}
\label{sec:anchorflow}

\begin{figure}[t]
    \centering
    \includegraphics[width=\columnwidth]{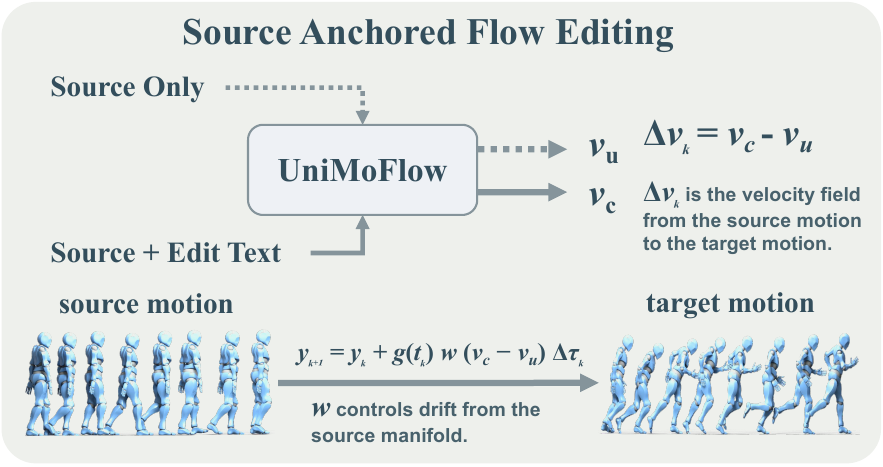}
    \vspace{-3pt}
    \caption{SAFE inference. Starting from the source latent, UniMoFlow evaluates source-conditioned velocity fields with and without the edit instruction. Their gated difference updates the trajectory, while drift strength $w$ controls the trade-off between source fidelity and edit magnitude.}
    \label{fig:safe-sampling}
    \vspace{-0.5em}
\end{figure}

The training paradigm above yields a controllable inference-time editing capability as a natural byproduct. UniMoFlow learns not only the instruction-conditioned velocity field $\mathbf{v}_\theta(\cdot, \mathbf{z}_s, c_e)$, but also a robust \emph{source-conditioned null-text} field $\mathbf{v}_\theta(\cdot, \mathbf{z}_s, \varnothing)$. The residual between these two fields distills the velocity component attributed to the edit instruction while remaining anchored to the source motion. We term this strategy SAFE, \emph{Source-Anchored Flow Editing}. Figure~\ref{fig:safe-sampling} summarizes this inference process: the instruction-specific velocity residual steers the source-initialized trajectory, while $w$ directly controls its drift.

SAFE is conceptually inspired by FlowEdit \citep{kulikov2025flowedit}, which performs editing by integrating the difference between two text-conditioned velocity fields:
\begin{equation}
    \Delta \mathbf{v}_{\mathrm{FE}}(\mathbf{z}, t) = {v}_\theta(\mathbf{z}, t, c_{\mathrm{tgt}}) - {v}_\theta(\mathbf{z}, t, c_{\mathrm{src}}),
\label{eq:flowedit}
\end{equation}
SAFE differs by explicitly using the source motion as both trajectory initialization and condition for both velocity fields. Therefore, source preservation is enforced by construction rather than only through prompt cancellation, and no source caption is required at inference.

Concretely, SAFE initializes the generative trajectory with the source latent, $\mathbf{y}_0 = \mathbf{z}_s$. At step $k$, corresponding to nominal time $\tau_k$, the current state is re-noised: $\mathbf{z}_t^{\tau_k} = (1-\tau_k)\mathbf{y}_k + \tau_k \boldsymbol{\epsilon}$. We then evaluate the instruction-conditioned and null-text velocities in editing mode:
\begin{equation}
    {v}_c = {v}_\theta(\mathbf{z}^{\tau_k}_t, \mathbf{z}_s, c_e, \tau_k),
    \,\,
    {v}_u = {v}_\theta(\mathbf{z}^{\tau_k}, \mathbf{z}_s, \varnothing, \tau_k).
\label{eq:af_velocities}
\end{equation}
Modulated by a guidance scale $w$ and a time-dependent gate $g(t_k) \in [0,1]$, the source-anchored update step is defined as:
\begin{equation}
    \mathbf{y}_{k+1} = \mathbf{y}_k + g(t_k)\, w\, ({v}_c - {v}_u)\, \Delta \tau_k .
\label{eq:af_update}
\end{equation}

\afterpage{
\begin{figure*}[!t]
\setlength{\dbltextfloatsep}{2pt plus 1pt minus 1pt}
\centering
\vspace{-0.25em}
    \includegraphics[width=\textwidth]{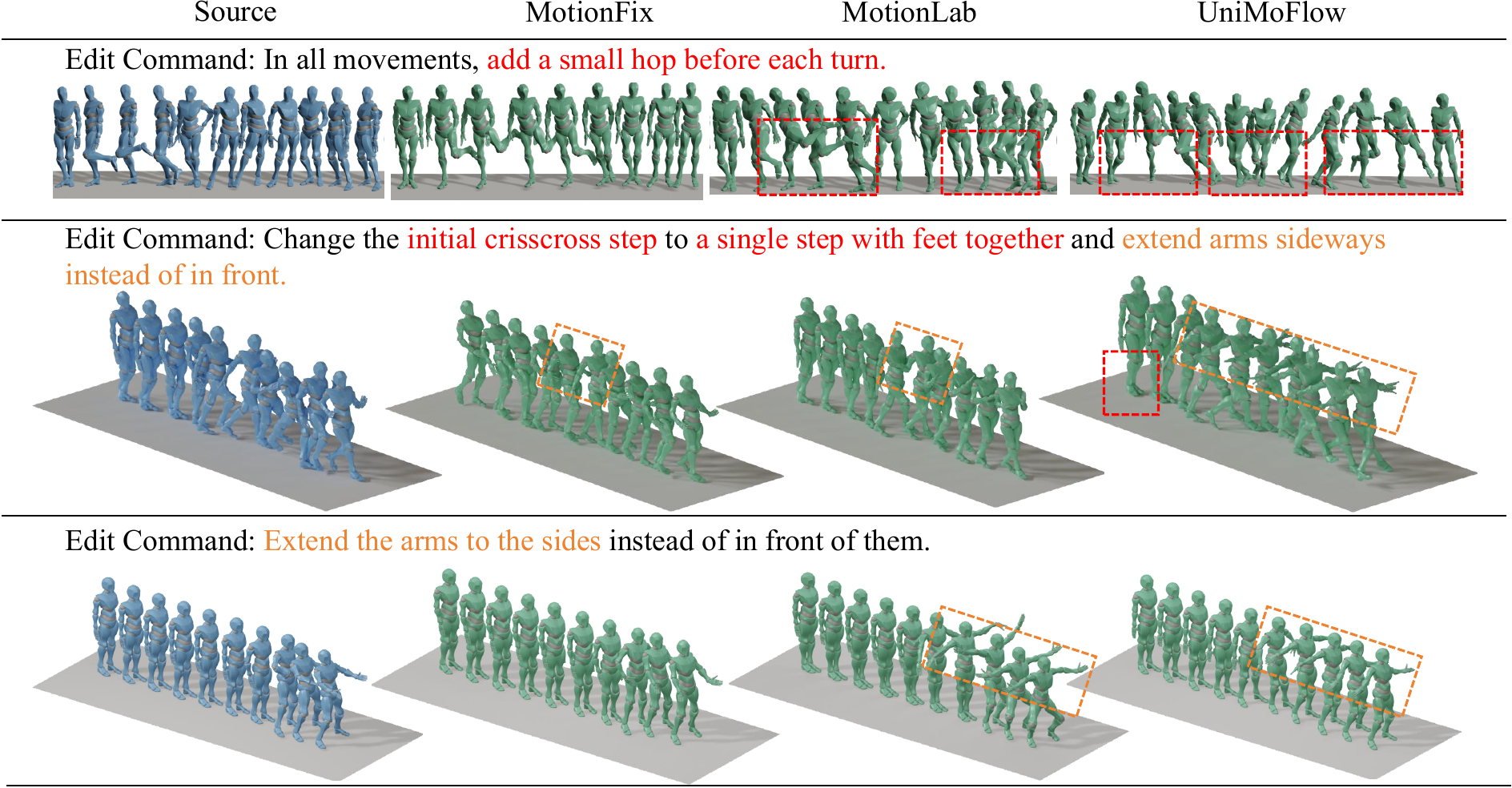}
    \caption{Qualitative comparison on Omni-MoEdit. \textcolor{red}{Red} and \textcolor{captionorange}{orange} boxes highlight motion regions that satisfy the edit-command semantics.}

    \label{fig:vis-compare}
\vspace{0pt}
\end{figure*}
}
The hyperparameters $(w, g)$ provide continuous control over editing strength. Smaller guidance values yield source-faithful refinements, while larger values induce stronger modifications. SAFE is therefore a complementary inference strategy rather than a replacement for native supervised editing: for drastic structural reconstructions, excessive guidance may drift off the plausible motion manifold, whereas insufficient guidance leaves the source under-edited. In such aggressive regimes, the native editor is more reliable. We compare these operating regimes in Section~\ref{sec:experiments}.

\section{Experiments}\label{sec:experiments}
We evaluate UniMoFlow on instruction-driven editing and text-to-motion generation, covering datasets, settings, metrics, and comparisons with representative baselines.

\subsection{Datasets and Experimental Setup}

Editing is evaluated on Omni-MoEdit, whose samples contain a source motion, edit instruction, target description, and filtered synthesized target motion. We compare with TMED, the editing model introduced by MotionFix, as well as MotionReFit, OmniME, MotionLab, and SimMotionEdit \citep{athanasiou2024motionfix,jiang2025motionrefit,shi2026omnime,guo2025motionlab,li2025simmotionedit}. UniMoFlow is tested in two modes: the native editor, UniMoFlow (Native), and the source-anchored sampler, UniMoFlow (SAFE), with step-size scaling fixed to 1.0.

For generation, we compare on the original SnapMoGen test split against T2M-GPT, MoMask, MoMask++, MDM, StableMoFusion, MARDM, Unified Flow, and our Base DiT synthesis backbone \citep{zhang2023generating,guo2024momask,guo2025snapmogen,tevet2023human,huang2024stablemofusion,meng2025rethinking,li2026unifiedflow}. Since Unified Flow is not open-source, we only report its published numbers and do not use it for data synthesis.

\textbf{Training configuration.} UniMoFlow uses 32-D Causal-VAE tokens with a 9-layer Transformer (width 1024, FFN 2048, 8 heads); UMT5-XXL uses length 150. Motions span 128--320 frames at 30 FPS and are downsampled by 4. We train for 500 epochs. Each epoch covers all generation and editing data; each step computes both losses and optimizes $\mathcal{L}=\lambda_g\mathcal{L}_{g}+\lambda_e\mathcal{L}_{e}$ with equal weights (batch 128). We use 0.5 reverse-instruction augmentation, AdamW, learning rate $2\times10^{-4}$, weight decay $10^{-5}$, 2000 warm-up iterations, and a MultiStep schedule at 100K/150K iterations. Evaluation uses 10 flow steps and CFG 4.5 on one A100 40GB GPU; full backbone and VAE settings are in Supplementary \emph{Model and Training Configuration}.

\subsection{Evaluation Metrics}
We evaluate editing from three complementary perspectives: target semantic alignment, source preservation, and actual edit effectiveness. For target alignment, TR@1/2/3 reports retrieval accuracy between target-text embeddings and edited-motion embeddings, while Match measures their embedding similarity. Struct measures source preservation as similarity between the source and edited motion in the evaluator pose space. Region checks whether the dominant temporal changes occur in segments semantically related to the instruction, and PosRatio reports how often editing improves the target-text matching score over the unedited source.

The MotionFix-style protocol reports edited-vs-GT R@3, edited-vs-source R@3, and FID for target fidelity, source retention, and target-distribution quality; Cycle-Con additionally measures reverse-edit recovery after a successful forward edit.

For text-to-motion generation, R-Precision Top-3 and Matching Score measure text-motion alignment, FID measures distribution quality, and MModality measures conditional diversity. All metric definitions and computation procedures are given in Supplementary \emph{Evaluation Protocol and Metrics}.

\subsection{Comparison with Baselines}
\setcounter{dbltopnumber}{3}
\renewcommand{\dblfloatpagefraction}{0.5}
\renewcommand{\dbltopfraction}{0.98}
\renewcommand{\textfraction}{0.02}
\setlength{\dblfloatsep}{2pt plus 1pt minus 1pt}
\setlength{\dbltextfloatsep}{2pt plus 1pt minus 1pt}

\textbf{Comparison scope.} For controlled comparison, we evaluate editors trained end-to-end from scratch on Omni-MoEdit. We exclude SOTA systems that fine-tune large open-source pretrained models, such as UMO~\citep{cong2026umo}, because inherited priors and fixed architectures are outside our focus and would confound evaluation of the compact unified generation--editing architecture developed here.

Table~\ref{tab:edit-target-text} shows that UniMoFlow (SAFE) achieves the best target-text retrieval, Match, and PosRatio, while UniMoFlow (Native) obtains the best Region score. SimMotionEdit and MotionReFit degrade because they compress the source motion into a weak auxiliary condition, which is insufficient for long, detail-rich Omni-MoEdit sequences.

Table~\ref{tab:edit-gt} shows that UniMoFlow (SAFE) leads in GT R@1, Cycle-Con, and FID; TMED preserves more source content but matches the target less accurately. Compared with MotionLab, the strongest non-UniMoFlow target-retrieval baseline, SAFE raises TR@1 from 0.4839 to 0.6347 and GT R@1 from 0.5197 to 0.6762, while reducing FID from 26.06 to 12.45. Complete GT/Src. R@1--R@3 results appear in Supplementary Table~\ref{tab:appendix-mix-ablation}. Together, the protocols show that source similarity alone does not imply successful editing: SAFE improves target compliance and cycle consistency while preserving distributional quality. This favors SAFE for localized changes and native editing for larger structural reconstruction. Table~\ref{tab:t2m-snapmogen} shows that UniMoFlow improves over Base DiT and obtains the best FID and Matching Score among continuous-space generators, while discrete-token models retain stronger retrieval. Mixed-training, architectural, latent-space, and CFG ablations are reported in Supplementary \emph{Ablation Studies}.

\subsection{Qualitative Results}

Figure~\ref{fig:vis-compare} compares UniMoFlow with TMED and MotionLab. In row 1, UniMoFlow preserves all three turns and inserts a hop before each, demonstrating repeated temporal grounding. In row 2, it satisfies the coupled footwork and arm-direction changes while retaining the crisscross-step structure; the baselines realize only part of the instruction or distort surrounding motion. In row 3, both UniMoFlow and MotionLab follow the target semantics, but UniMoFlow produces a smoother, more source-faithful result. Across seven additional test cases in Supplementary Figure~\ref{fig:supplementary-comparison}, UniMoFlow more consistently realizes multiple color-coded clauses in their corresponding temporal and body regions. Supplementary Figure~\ref{fig:five-basic-edits} further shows that action type, body part, amplitude, timing, and style can each be controlled from the same source. Together, these examples indicate that full source-motion context supports precise edits while retaining unspecified content. This indicates localized rather than global regeneration.

\section{Conclusion}

We presented Omni-MoEdit and UniMoFlow for instruction-driven 3D human motion generation and editing. Omni-MoEdit scales edit-pair construction with Qwen3-8B-generated instructions, flow-matching target synthesis, automatic filtering, source/target captions, and reverse instructions. UniMoFlow unifies generation and editing as two modes of a continuous latent flow-matching model with mask-controlled global self-attention over text, noisy target motion, and full source-motion tokens. Synthesized edit data teaches native editing behavior, while high-quality text-motion data regularizes the shared generative prior. SAFE further provides controllable source-faithful sampling.

Experiments show that UniMoFlow improves semantic editing, target-motion alignment, and cycle consistency, and achieves the best FID and CLIP-based matching score among continuous-space generation methods. The system still depends on automatic filtering and may struggle with edits requiring substantial source-motion reconstruction. Future work will explore stronger data validation, human preference signals, and adaptive sampling.

\clearpage

\clearpage
\bibliography{aaai2027}

\clearpage
\appendix
\twocolumn[
\begin{center}
{\LARGE\bfseries Supplementary}
\end{center}
\vspace{0.4em}
]

\setlength{\textfloatsep}{6pt plus 2pt minus 2pt}
\setlength{\dbltextfloatsep}{5pt plus 2pt minus 1pt}
\setlength{\floatsep}{5pt plus 2pt minus 1pt}
\setlength{\dblfloatsep}{4pt plus 2pt minus 1pt}
\setlength{\intextsep}{6pt plus 2pt minus 2pt}

\section{Overview}
This supplementary material is organized into six parts. \emph{Evaluation Protocol and Metrics} defines the complete editing and generation evaluation protocol. \emph{Omni-MoEdit Dataset Details} details the synthesis schema, split isolation, scale, and edit composition. \emph{Model and Training Configuration} records the Base DiT and 1D Wan VAE settings. \emph{Reproduction of Open-Source Editors} specifies the common data protocol, method-specific adaptations, training hyperparameters, checkpoints, and sampling settings for all editing baselines. \emph{Ablation Studies} reports mixed-training, architectural, latent-space, and guidance ablations. \emph{Additional Qualitative Results} provides broader comparisons, examples of the five basic edit types, and representative failure cases.

\section{Evaluation Protocol and Metrics}
\label{app:metrics}
This section provides the definitions and computation details of the evaluation protocol used in the main paper.

\subsection{Metric Definitions}
This section gives the explicit definitions used by the evaluator. Let $\mathbf{x}_{s,i}$, $\hat{\mathbf{x}}_i$, $\mathbf{x}_{t,i}$, and $\mathbf{x}_{cyc,i}$ denote the source motion, edited motion, synthesized target motion, and reverse-edited motion for sample $i$, respectively. Let $c_{s,i}$, $c_{t,i}$, and $c_{e,i}$ denote the source caption, target caption, and edit instruction. A frozen SnapMoGen evaluator maps motions and texts into a shared embedding space, denoted by $\phi_M(\cdot)$ and $\phi_T(\cdot)$. All metrics are averaged over valid test samples.

\paragraph{Retrieval and matching.}
For paired query embeddings $\mathbf{q}_i$ and candidate embeddings $\mathbf{z}_i$, we define the rank of the matched candidate as
\begin{equation}
    \rho_i = 1 + \sum_{j\ne i}\mathbb{I}\left[d(\mathbf{q}_i,\mathbf{z}_j)<d(\mathbf{q}_i,\mathbf{z}_i)\right],
\end{equation}
where $d(\cdot,\cdot)$ is the evaluator distance. The retrieval metric is
\begin{equation}
    \mathrm{R@}k = \frac{1}{N}\sum_{i=1}^{N}\mathbb{I}[\rho_i\le k].
\end{equation}
For target-text-centered editing evaluation, TR@$k$ uses $\mathbf{q}_i=\phi_T(c_{t,i})$ and $\mathbf{z}_i=\phi_M(\hat{\mathbf{x}}_i)$. The target matching score is
\begin{equation}
    \mathrm{Match}=\frac{1}{N}\sum_{i=1}^{N}\cos\left(\phi_M(\hat{\mathbf{x}}_i),\phi_T(c_{t,i})\right).
\end{equation}
For MotionFix-style reference evaluation, Edited-vs-GT retrieval uses $\mathbf{q}_i=\phi_M(\hat{\mathbf{x}}_i)$ and $\mathbf{z}_i=\phi_M(\mathbf{x}_{t,i})$, while Edited-vs-Source retrieval uses $\mathbf{z}_i=\phi_M(\mathbf{x}_{s,i})$.

\paragraph{Positive improvement ratio.}
Target retrieval alone may reward an almost unchanged source motion if the source already partially matches the target caption. We therefore compute the target-alignment gain of each edit:
\begin{align}
    S_i^{src} &= \cos\left(\phi_M(\mathbf{x}_{s,i}),\phi_T(c_{t,i})\right),\\
    S_i^{edit} &= \cos\left(\phi_M(\hat{\mathbf{x}}_i),\phi_T(c_{t,i})\right),\\
    \Delta S_i &= S_i^{edit}-S_i^{src}.
\end{align}
The positive improvement ratio is
\begin{equation}
    \mathrm{PosRatio}=\frac{1}{N}\sum_{i=1}^{N}\mathbb{I}[\Delta S_i>0].
\end{equation}
Its design goal is to verify that the output moves toward the requested target semantics relative to the source, rather than only preserving a plausible source structure.

\paragraph{Structure preservation.}
We measure global source preservation in pose space before computing localized edit metrics. Following the implementation, only the first 148 motion channels are used. With $P(\cdot)$ selecting these channels and $\operatorname{vec}(\cdot)$ flattening the valid temporal range, the per-sample score is
\begin{equation}
    \mathrm{Struct}_i = \cos\left(\operatorname{vec}(P(\mathbf{x}_{s,i}^{1:L_i})),\operatorname{vec}(P(\hat{\mathbf{x}}_i^{1:L_i}))\right).
\end{equation}
Struct is high when the edited motion preserves the overall source trajectory and body configuration, and low when the edit drifts away from the source.

\paragraph{Region alignment.}
Region is introduced to estimate whether the most changed temporal region is actually relevant to the edit instruction. First, the evaluator computes frame-wise edit magnitude
\begin{equation}
    \delta_{i,t}=\left\|\hat{\mathbf{x}}_{i,t}-\mathbf{x}_{s,i,t}\right\|_2,\quad t=1,\ldots,L_i.
\end{equation}
Let $q_i=Q_{75}(\{\delta_{i,t}\}_{t=1}^{L_i})$ be the 75th percentile of motion changes. Candidate edit frames are selected by
\begin{equation}
    m_{i,t}=\mathbb{I}[\delta_{i,t}>q_i].
\end{equation}
The positive frames in $m_i$ are grouped into contiguous segments, and the longest segment is selected. If the segment is shorter than eight frames, it is symmetrically expanded when valid frames are available. Denote the final segment by $[a_i,b_i)$. The Region score is then
\begin{equation}
    \mathrm{Region}_i=\cos\left(\phi_M(P(\hat{\mathbf{x}}_i^{a_i:b_i})),\phi_T(c_{e,i})\right).
\end{equation}
This construction separates the magnitude of change from the semantics of change: a method receives a high Region score only when its strongest local modification is aligned with the edit command. The implementation also computes the selected-frame coverage
\begin{equation}
    \mathrm{Coverage}_i=\frac{1}{L_i}\sum_{t=1}^{L_i}m_{i,t},
\end{equation}
and the concentration ratio
\begin{equation}
    \mathrm{PeakRatio}_i=\frac{\max_t\delta_{i,t}}{\frac{1}{L_i}\sum_{t=1}^{L_i}\delta_{i,t}+\epsilon},
\end{equation}
which help diagnose whether an edit is too diffuse or dominated by a small temporal spike.

\paragraph{Cycle consistency.}
Cycle-Con is designed for datasets that include reverse instructions. Given a forward edit $\mathbf{x}_{s,i}\rightarrow\hat{\mathbf{x}}_i$, we apply the reverse instruction to produce $\mathbf{x}_{cyc,i}$ and measure whether it returns to the original source. The source-return structure term is
\begin{equation}
    C_i^{struct}=\cos\left(\operatorname{vec}(P(\mathbf{x}_{s,i}^{1:L_i})),\operatorname{vec}(P(\mathbf{x}_{cyc,i}^{1:L_i}))\right).
\end{equation}
We also compute a text-based return term and a motion-based return term:
\begin{align}
    C_i^{text} &= \mathrm{R@1}\left(\phi_T(c_{s,i}),\phi_M(\mathbf{x}_{cyc,i})\right),\\
    C_i^{motion} &= \mathrm{R@1}\left(\phi_M(\mathbf{x}_{cyc,i}),\phi_M(\mathbf{x}_{s,i})\right).
\end{align}
The ungated return score used in the implementation is
\begin{equation}
    R_i^{cyc}=0.40C_i^{struct}+0.30C_i^{text}+0.30C_i^{motion}.
\end{equation}
A pure copy of the source could be easy to cycle back but would not be a valid edit. We therefore gate the return score by the forward semantic gain and by the amount of non-trivial change:
\begin{align}
    G_i^{sem} &= \operatorname{clip}(\Delta S_i/0.05,0,1),\\
    G_i^{chg} &= \operatorname{clip}((1-\mathrm{Struct}_i)/0.05,0,1),\\
    G_i &= \sqrt{G_i^{sem}G_i^{chg}}.
\end{align}
The final cycle metric is
\begin{equation}
    \mathrm{CycleCon}_i=R_i^{cyc}G_i.
\end{equation}
Cycle-Con therefore favors edits that are effective in the forward direction, preserve enough source information to be reversible, and can be reconstructed by the reverse instruction.

\paragraph{FID and multimodality.}
FID is computed in the frozen evaluator motion-embedding space. Given generated embeddings with mean and covariance $(\mu_g,\Sigma_g)$ and reference embeddings $(\mu_r,\Sigma_r)$,
\begin{equation}
    \mathrm{FID}=\|\mu_g-\mu_r\|_2^2+\operatorname{Tr}\left(\Sigma_g+\Sigma_r-2(\Sigma_g\Sigma_r)^{1/2}\right).
\end{equation}
For text-to-motion generation, MModality measures diversity by sampling multiple motions for the same text and averaging pairwise distances between their evaluator embeddings:
\begin{equation}
    \mathrm{MModality}=\frac{1}{N}\sum_{i=1}^{N}\frac{1}{|\mathcal{P}_i|}\sum_{(a,b)\in\mathcal{P}_i}\left\|\phi_M(\mathbf{x}_{i}^{a})-\phi_M(\mathbf{x}_{i}^{b})\right\|_2.
\end{equation}

\section{Omni-MoEdit Dataset Details}
\label{app:dataset}
This section summarizes the synthesis protocol and composition of Omni-MoEdit.

\subsection{Synthesis Prompt and Split Protocol}
For each SnapMoGen source caption, Qwen3-8B generates $N$ structured candidates containing a forward edit instruction, a minimally revised target caption, and a reverse instruction. The prompt requires the instruction to specify the affected timing, body part, modification type, and intended style whenever applicable; it discourages ambiguous commands and asks the target caption to preserve unchanged source semantics. The base text-to-motion DiT then synthesizes the target motion with FlowEdit, after which the automatic filters described in the main paper retain candidates with sufficient target alignment, semantic improvement over the source, and motion structure.

Each Omni-MoEdit split is constructed only from the corresponding SnapMoGen split: training sources produce training edits, validation sources produce validation edits, and test sources produce test edits. This prevents source-motion leakage across splits. Each accepted record contains source and target motions, source and target captions, a forward instruction, and a reverse instruction. The resulting 55,641 pairs comprise 46,911/2,895/5,835 train/validation/test examples from 22,941 unique sources and 44,836 synthesized targets.

\subsection{Dataset Composition}
The five edit categories overlap because one instruction may modify several attributes. Body-part edits occur in 46,024 pairs (82.7\%), followed by timing (26,357; 47.4\%), action type (22,058; 39.6\%), amplitude (19,567; 35.2\%), and style (13,020; 23.4\%). Only 148 pairs fall outside these categories, whereas 46,516 pairs (83.6\%) combine at least two types. The most frequent exact combinations are action type plus body part (6,575), body part only (5,800), amplitude plus body part (5,184), body part plus timing (5,178), and amplitude--body-part--timing (5,134).

Target-caption analysis further shows broad motion coverage, including gestures, posing, gaze changes, turns, locomotion, bending, sitting, jumping, dancing, object interaction, running, striking, squatting, throwing/catching, crawling, and falling. Only 28 target captions are unmatched by this descriptive taxonomy. These rule-based categories are used only for analysis, not as benchmark labels.

\begin{figure*}[t]
\centering
\includegraphics[width=\textwidth]{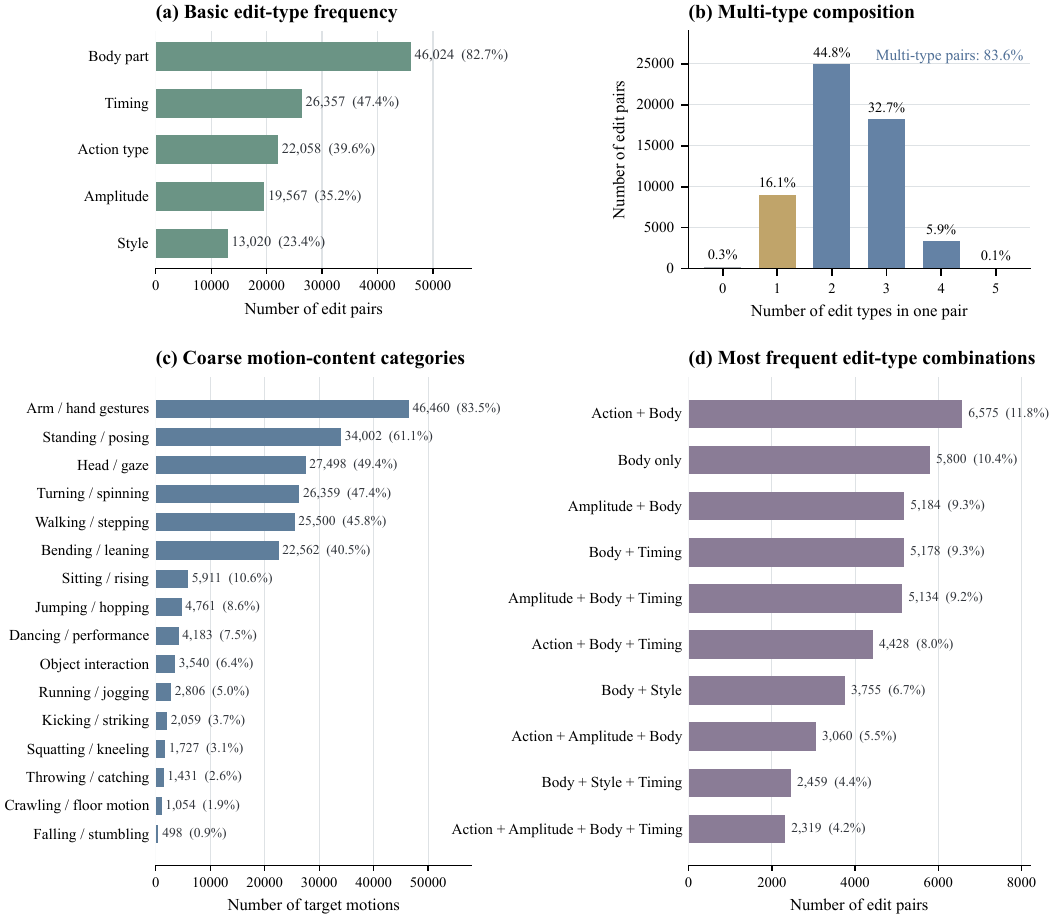}
\caption{Composition of Omni-MoEdit. (a) Overlapping frequencies of the five basic edit types. (b) Number of edit types assigned to each pair. (c) Multi-label motion-content categories extracted from target captions. (d) The ten most frequent exact edit-type combinations. Frequencies in (a) and (c) overlap and therefore do not sum to 100\%.}
\label{fig:omni-moedit-statistics}
\end{figure*}

\section{Model and Training Configuration}
\label{app:configuration}
This section records the principal baseline, representation, and reproducibility settings used by the experiments.

\subsection{Base DiT and 1D Wan VAE}
The Base DiT baseline in Table~\ref{tab:t2m-snapmogen} is a text-to-motion latent flow-matching DiT. It uses a modified 1D Wan Transformer denoiser with hidden dimension 1024, FFN dimension 2048, 9 Transformer blocks, 8 attention heads, and velocity prediction. Text is encoded by UMT5-XXL with text dimension 4096 and maximum text length 150. Each block first applies RoPE-based self-attention over motion latent tokens and then injects the projected text memory through a separate cross-attention module before the FFN. The baseline uses classifier-free guidance with training text dropout 0.1 and inference CFG scale 5.0, 10 sampling steps, batch size 32, learning rate $2\times10^{-4}$, weight decay 0, and 2,000 warm-up iterations. The denoiser contains approximately 126.4M trainable parameters. This baseline isolates the effect of replacing cross-attention text injection with UniMoFlow's unified in-context token sequence and mask-controlled global self-attention.

Both Base DiT and UniMoFlow operate on the same 1D Wan VAE latent space. The VAE adapts the causal VAE design of Wan-style video generation to 1D motion features: input motions are represented by 296-dimensional frame-wise features at 30 FPS, processed by causal 1D convolutions, RMS normalization, residual blocks, and temporal down/up-sampling modules. The encoder width is 384, the decoder width is 768, the latent dimension is 32, and the temporal down-sampling pattern is $[\mathrm{True},\mathrm{True},\mathrm{False}]$, yielding a $4\times$ temporal compression. Thus every four motion frames are compressed into one latent token while preserving temporal causality. The VAE is trained with Smooth-L1 reconstruction losses in feature, joint, and velocity spaces, plus a small KL regularization term; the training uses batch size 512, learning rate $3\times10^{-4}$, cosine scheduling, EMA, and gradient clipping. The VAE contains approximately 42.7M trainable parameters.

\section{Reproduction of Open-Source Editors}
\label{app:baseline-reproduction}

\subsection{Common Protocol}
We reproduce TMED, MotionReFit, SimMotionEdit, MotionLab, and OmniME from their released implementations~\citep{athanasiou2024motionfix,jiang2025motionrefit,li2025simmotionedit,guo2025motionlab,shi2026omnime}. Only the data loaders, dimensional adapters, and evaluation hooks are changed. Every model uses the same 46,911/2,895/5,835 Omni-MoEdit train/validation/test partition and the forward source--instruction--target triplets; validation and test motions never enter training. The native motions are 296-dimensional features at 30 FPS and are standardized with statistics computed from the training split. Variable-length sequences are padded only within each batch unless a method requires a fixed temporal interface. We do not add MotionFix, STANCE, MotionCutMix, or MotionLab's auxiliary task corpora, so the comparison isolates how each released editor learns from Omni-MoEdit. All runs use one NVIDIA A100 40GB GPU. Hydra-based runs use seed 42; MotionReFit and MotionLab use seed 1234.

\subsection{Method-Specific Settings}
\paragraph{TMED.}
TMED is the conditional diffusion editor introduced with MotionFix~\citep{athanasiou2024motionfix}. Our reproduction keeps its compressed source-motion conditioning and frozen T5-v1.1-base text encoder. The denoiser has eight Transformer blocks, width 512, FFN width 1024, four attention heads, and dropout 0.1. It predicts the clean sample under a 300-step cosine diffusion process. We use AdamW with batch size 128, learning rate $10^{-4}$, 150 warm-up steps, and final learning rate $10^{-6}$. Text and motion conditions are each dropped with probability 0.05 for classifier-free guidance. The reported checkpoint is sampled for 200 steps with text/motion guidance scales 2.5/2.0.

\paragraph{MotionReFit.}
MotionReFit is an autoregressive conditional diffusion editor that processes motion in short windows and uses a motion coordinator~\citep{jiang2025motionrefit}. Following its native interface, we convert each Omni-MoEdit frame to 28 three-dimensional joints and train on aligned 16-frame windows sampled with frame stride two; the first two frames provide fixed autoregressive context. The frozen text encoder is CLIP ViT-B/32. The denoiser has eight layers, width 512, 16 heads, and dropout 0.1, with 100 linear diffusion steps and condition-drop probability 0.1. We use AdamW, batch size 128, learning rate $10^{-4}$, and gradient clipping at 1.0. To keep the training data identical across baselines, we disable additional MotionCutMix/STANCE augmentation and train only on Omni-MoEdit.

\paragraph{SimMotionEdit.}
SimMotionEdit originally augments editing with motion-similarity prediction~\citep{li2025simmotionedit}. Its released auxiliary labels are tied to MotionFix-specific source--target preprocessing and are not available for Omni-MoEdit. The reproducible comparison therefore retains the released source/text diffusion path and Transformer source encoder, but disables the auxiliary similarity head rather than importing extra MotionFix supervision. The eight-block denoiser uses width 512, FFN width 1024, four heads, dropout 0.1, and a frozen T5-v1.1-base encoder. Training uses a 1,000-step cosine process, condition-drop probabilities of 0.2, AdamW with batch size 128 and learning rate $10^{-4}$, 150 warm-up steps, and final learning rate $10^{-6}$. The reported checkpoint uses 50-step DDIM sampling. Thus this row measures the open-source SimMotionEdit editing path under the same triplet-only supervision as the other baselines.

\paragraph{MotionLab.}
MotionLab formulates multiple generation and editing tasks as source motion--condition--target motion rectified flow~\citep{guo2025motionlab}. We instantiate its \texttt{source\_text} specialist only, without auxiliary datasets or curriculum tasks. It operates directly on 296-dimensional motions, randomly crops training sequences to at most 320 frames, and pads shorter sequences by repeating the last frame; no additional VAE is used. The MotionFlow denoiser has nine blocks, token width 512, FFN width 1024, eight heads, and dropout 0.1. A frozen T5-v1.1-base encoder uses maximum length 77. Source+text, source-only, and unconditional training probabilities are 0.8/0.1/0.1. We train with AdamW, batch size 64, learning rate $10^{-4}$, and gradient clipping at 1.0. The reported checkpoint uses 50 Euler flow steps.

\paragraph{OmniME.}
OmniME balances change and preservation through positive--negative supervision~\citep{shi2026omnime}. We preserve its three-level similarity classification and negative-text branches, while constructing the required annotations from Omni-MoEdit itself: source--target frame distances produce normalized per-frame similarity scores, a motion-SNR test masks unreliable pairs, and a command from a different edit type supplies the negative text. The local setting uses velocity-aware class weights, three classes, classification coefficient 0.05, and zero coefficients for the optional identity and source/target representation losses. Its eight-block DiT uses width 512, FFN width 1024, eight heads, a four-layer retrospective encoder, dropout 0.1, and frozen T5-v1.1-base text features. We use a 1,000-step cosine diffusion process, condition-drop probabilities of 0.2, AdamW with batch size 64 and learning rate $10^{-4}$, and 150 warm-up steps. The  checkpoint is evaluated with 50 denoising steps.

\subsection{Checkpoint and Evaluation Protocol}
The training launchers allow at most 1,001 epochs, while the reported runs are the final converged checkpoints listed above. Validation is run on the complete validation split every five epochs (every ten for SimMotionEdit), and all reported comparisons use the complete 5,835-pair test split with one sample per pair and the common evaluator defined in Section~\ref{app:metrics}. Sampling follows each method's standard editing path; debugging-only source-noise variants are excluded. The released reproduction package will include the dataset adapters, launch commands, resolved configurations, checkpoints, and metric aggregation scripts.

\begin{table*}[!t]
\centering
\small
\renewcommand{\arraystretch}{1.08}
\setlength{\tabcolsep}{3.0pt}
\resizebox{\textwidth}{!}{
\begin{tabular}{lcccccccccc}
\toprule
& \multicolumn{4}{c}{Target-text} & \multicolumn{3}{c}{Edited vs. GT} & \multicolumn{3}{c}{Edited vs. Source} \\
\cmidrule(lr){2-5} \cmidrule(lr){6-8} \cmidrule(lr){9-11}
Setting & TR@1 $\uparrow$ & TR@2 $\uparrow$ & TR@3 $\uparrow$ & Match $\uparrow$ & R@1 $\uparrow$ & R@2 $\uparrow$ & R@3 $\uparrow$ & R@1 $\uparrow$ & R@2 $\uparrow$ & R@3 $\uparrow$ \\
\midrule
\multicolumn{11}{l}{\textbf{(a) Mixed generation--editing training ablation}} \\
GT & 0.8973 & 0.9806 & 0.9953 & 0.7696 & -- & -- & -- & -- & -- & -- \\
E+G SAFE & \textbf{0.6347} & \textbf{0.7972} & \textbf{0.8651} & \textbf{0.5861} & \textbf{0.6762} & \textbf{0.8263} & \textbf{0.8923} & \textbf{0.5909} & \textbf{0.8353} & \textbf{0.9359} \\
E+G Native & 0.5492 & 0.7036 & 0.7777 & 0.5263 & 0.5852 & 0.7302 & 0.8005 & 0.3664 & 0.5741 & 0.6985 \\
Edit only SAFE & \underline{0.6089} & \underline{0.7662} & \underline{0.8426} & \underline{0.5647} & \underline{0.6486} & \underline{0.8044} & \underline{0.8699} & \underline{0.5660} & \underline{0.8063} & \underline{0.9131} \\
Edit only Native & 0.5537 & 0.7021 & 0.7845 & 0.5229 & 0.5809 & 0.7348 & 0.8039 & 0.4207 & 0.6238 & 0.7424 \\
\midrule
\multicolumn{11}{l}{\textbf{(b) Base model architecture ablation (edit-only training)}} \\
UniMoFlow (Native) & \textbf{0.5537} & \underline{0.7021} & \textbf{0.7845} & \textbf{0.5229} & \textbf{0.5809} & \underline{0.7348} & \underline{0.8039} & 0.4207 & 0.6238 & 0.7424 \\
w/o Task Tag \& Task Head & \underline{0.5501} & \textbf{0.7103} & \underline{0.7798} & \underline{0.5223} & \underline{0.5774} & \textbf{0.7367} & 0.8001 & 0.4211 & 0.6199 & 0.7542 \\
w/o Asymmetric Attention & 0.5407 & 0.6938 & 0.7697 & 0.5108 & 0.5705 & 0.7244 & \textbf{0.8073} & 0.4519 & 0.6483 & 0.7835 \\
w/o Token-Level Tags & 0.5129 & 0.6791 & 0.7592 & 0.4985 & 0.5513 & 0.7006 & 0.7882 & \underline{0.4726} & \underline{0.6591} & \underline{0.8026} \\
w/o Latent Flow Matching (TMED) & 0.4262 & 0.6243 & 0.7362 & 0.4552 & 0.4579 & 0.6735 & 0.7835 & \textbf{0.6166} & \textbf{0.8580} & \textbf{0.9547} \\
\bottomrule
\end{tabular}}
\caption{Unified ablation study. Panel (a) evaluates mixed generation--editing training, where E+G denotes joint edit-generation training. Panel (b) evaluates the principal UniMoFlow modules under edit-only training; its final row is the TMED-style baseline without latent flow matching. GT and Source columns report edited-motion retrieval against the synthesized target and source motions. Bold and underline denote the best and second-best results within each panel, excluding the GT reference row.}
\label{tab:appendix-mix-ablation}
\label{tab:architecture-ablation}
\end{table*}

\begin{table}[!t]
\centering
\small
\renewcommand{\arraystretch}{1.05}
\setlength{\tabcolsep}{5pt}
\begin{tabular}{lccc}
\toprule
Latent dim. & g-FID $\downarrow$ & r-FID $\downarrow$ & MPJPE $\downarrow$ \\
\midrule
32 & \textbf{15.884} & \textbf{1.6837} & \textbf{4.3015} \\
24 & 18.922 & 4.3871 & 5.6172 \\
16 & 26.788 & 14.9988 & 7.4039 \\
\bottomrule
\end{tabular}
\caption{Ablation of the 1D Wan VAE latent dimension. g-FID is measured using Base DiT generation in the corresponding latent space; r-FID and MPJPE evaluate VAE reconstructions. Bold denotes the best result.}
\label{tab:vae-latent-dim}
\end{table}

\begin{figure}[!t]
\centering
\includegraphics[width=\columnwidth]{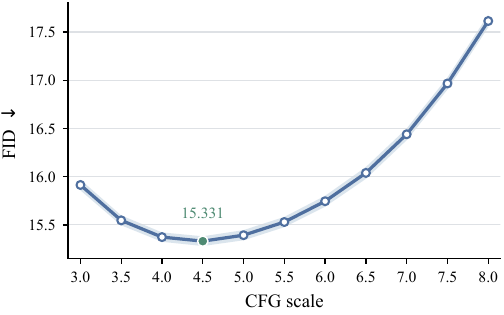}
\caption{Generation FID under different classifier-free guidance scales on the SnapMoGen test split. The shaded band shows the 95\% confidence interval over 20 repeated evaluations; lower is better.}
\label{fig:cfg-fid-ablation}
\end{figure}

\begin{figure*}[!t]
\centering
\includegraphics[width=\textwidth]{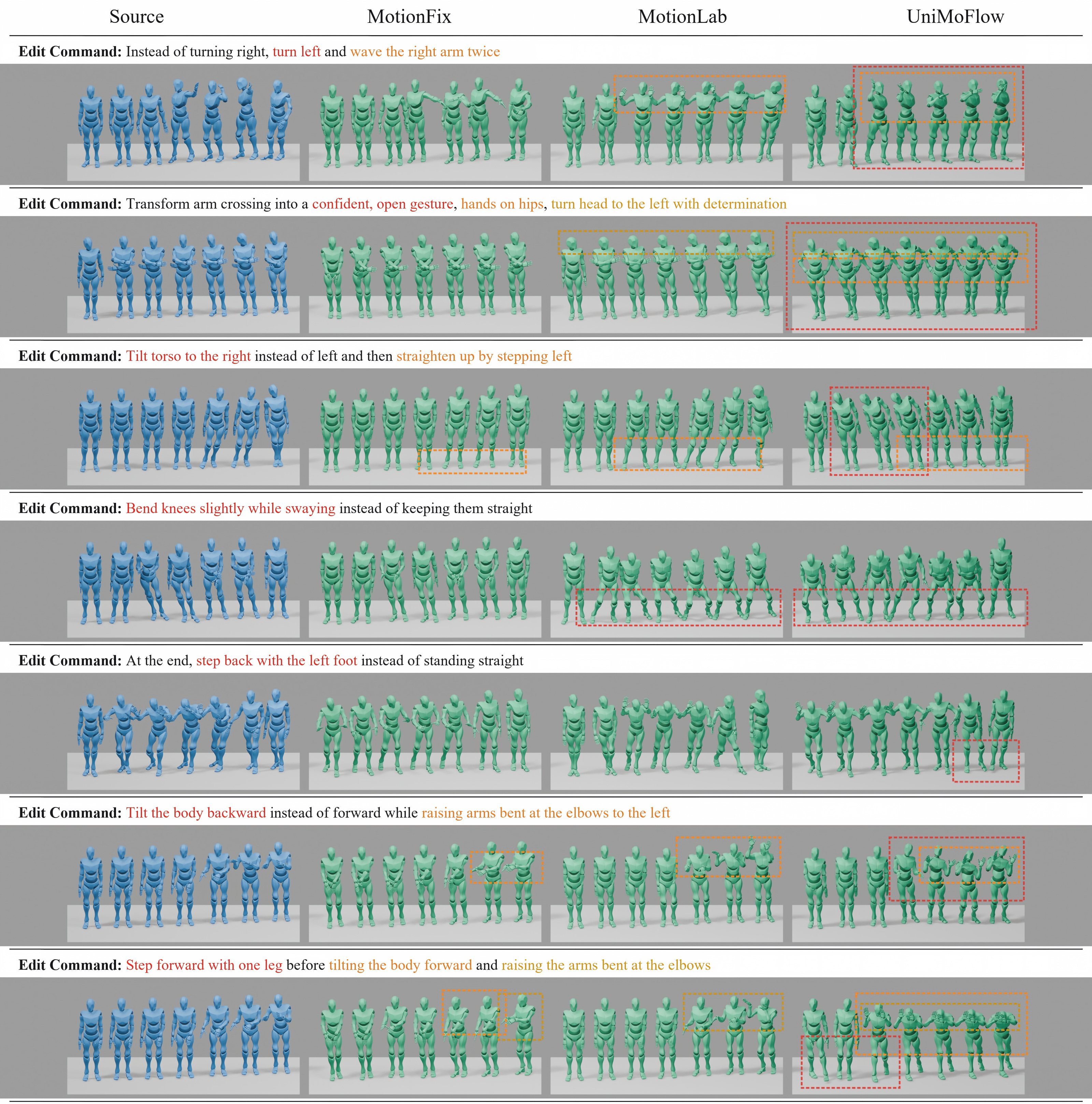}
\caption{Additional qualitative comparisons on the Omni-MoEdit test split. Each row applies the same source motion and edit command to MotionFix, MotionLab, and UniMoFlow. Red and orange dashed boxes localize the motion changes induced by the correspondingly colored instruction phrases; these boxes are visualization annotations rather than model inputs.}
\label{fig:supplementary-comparison}
\end{figure*}

\section{Ablation Studies}
\label{app:ablations}
Table~\ref{tab:appendix-mix-ablation} consolidates mixed-training and architecture ablations under a shared evaluation protocol.

\subsection{Mixed Generation--Editing Training}
Synthetic edit triplets alone already provide usable native editing ability. Table~\ref{tab:appendix-mix-ablation}(a) shows that mixing the same edit supervision with high-quality text-to-motion data improves target-text retrieval, matching, and source/target reference retrieval. Generation batches regularize residual artifacts in synthetic edits, while edit batches strengthen language--motion grounding, supporting the mutually beneficial training design.

\subsection{UniMoFlow Architecture}
All variants in Table~\ref{tab:architecture-ablation}(b) use editing data only. UniMoFlow is derived from TMED, the compact MotionFix editor~\citep{athanasiou2024motionfix}. Moving it into the causal-VAE latent space yields the largest gain: $4\times$ compression reduces a 320-frame sequence to at most 80 tokens, and latent flow matching raises TR@1 from 0.4262 to 0.5537 and Match from 0.4552 to 0.5229. Token-level tags and asymmetric attention add further target-alignment gains and, more importantly, enable generation and editing to share one backbone. The high source retrieval of the TMED-style row primarily reflects under-editing.

\subsection{VAE Latent Dimension}
Table~\ref{tab:vae-latent-dim} varies the latent channel dimension while holding the VAE architecture and $4\times$ temporal compression fixed. r-FID and MPJPE evaluate reconstruction, while g-FID uses Base-DiT generation in the corresponding latent space. Reducing the dimension consistently degrades both reconstruction and downstream generation; the 32-D latent is therefore used throughout.

\subsection{Classifier-Free Guidance}
Figure~\ref{fig:cfg-fid-ablation} evaluates CFG scales from 3.0 to 8.0 with 10 flow steps and 20 repeated evaluations. FID reaches its minimum of 15.331 at CFG 4.5 and then degrades to 17.613 at CFG 8.0; all main generation results therefore use CFG 4.5.

\subsection{SAFE Drift Strength}
\begin{table}[!t]
\centering
\small
\setlength{\tabcolsep}{5pt}
\begin{tabular}{c|cccc}
\toprule
Strength $w$ & Match $\uparrow$ & FID $\downarrow$ & TR@1 $\uparrow$ & GT R@1 $\uparrow$ \\
\midrule
0.5 & 0.5768 & 24.17 & 0.6144 & 0.6278 \\
\textbf{1.0} & \textbf{0.5850} & \textbf{12.45} & \textbf{0.6300} & \textbf{0.6692} \\
1.5 & 0.5721 & 13.40 & 0.6250 & 0.6629 \\
2.0 & 0.5542 & 20.13 & 0.6055 & 0.6468 \\
2.5 & 0.5379 & 29.93 & 0.5846 & 0.6324 \\
\bottomrule
\end{tabular}
\caption{SAFE editing under different drift strengths on Omni-MoEdit. Bold denotes the best setting.}
\label{tab:safe-strength-sweep}
\end{table}

SAFE multiplies the source-anchored conditional velocity difference by a drift strength $w$, which controls how far each sampling step moves from the source trajectory toward the instructed target. As shown in Table~\ref{tab:safe-strength-sweep}, $w=1.0$ applies the unscaled drift and simultaneously gives the highest Matching Score, TR@1, and GT R@1, together with the lowest FID. We therefore fix $w=1.0$ for the main-paper comparisons. Increasing $w$ beyond 1.0 progressively over-amplifies the velocity displacement: the generated motion moves away from the expected edited distribution, causing semantic alignment and retrieval to decline while FID rises. In interactive use, however, $w$ remains a useful continuous control and can be adjusted moderately to fine-tune the magnitude of a desired motion change.

\begin{figure*}[!t]\centering\includegraphics[width=0.80\textwidth]{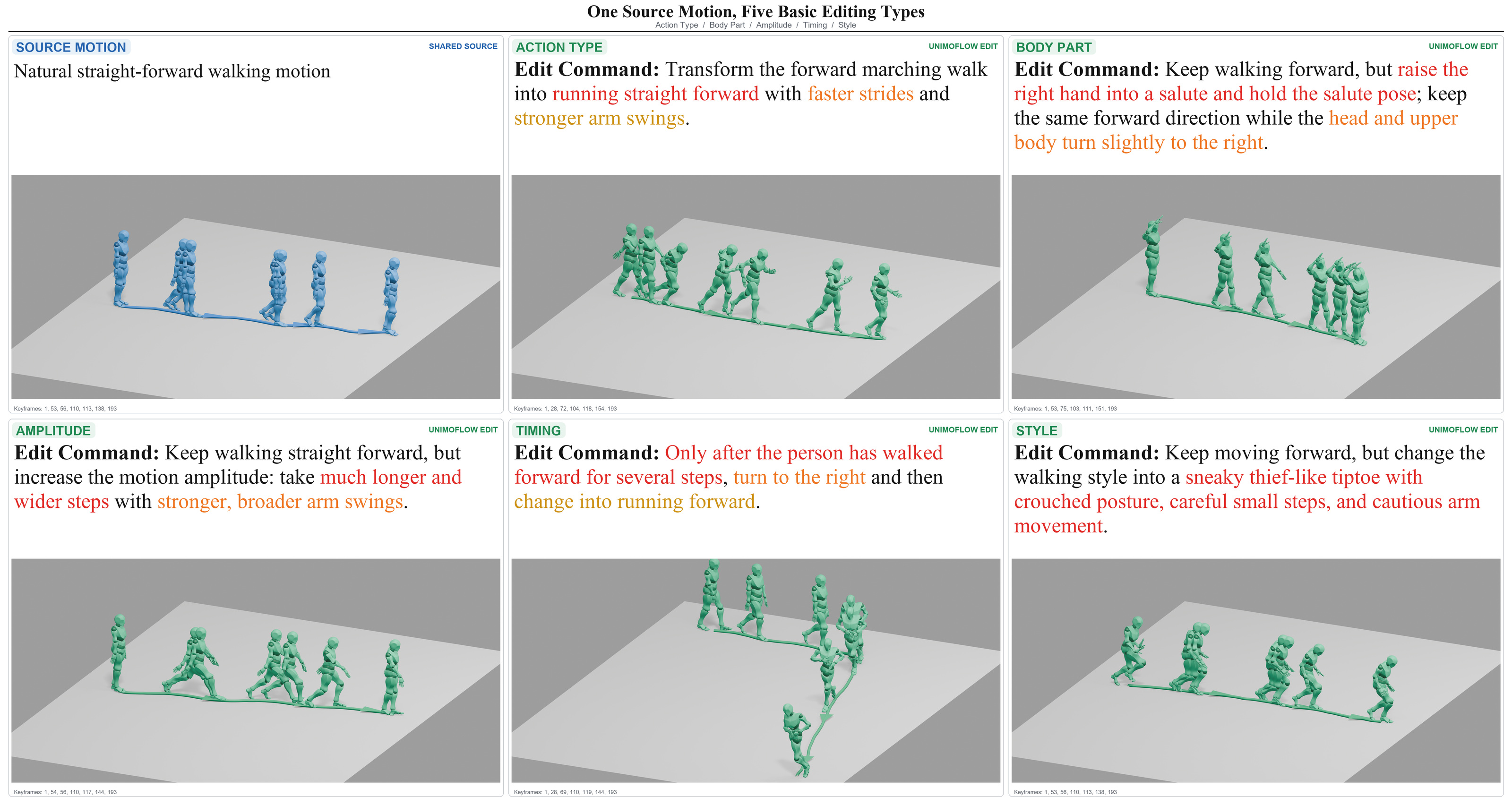}\caption{Five isolated UniMoFlow edits from one forward-walking source, covering action type, body part, amplitude, timing, and style.}\label{fig:five-basic-edits}\vspace{2pt}\includegraphics[width=0.80\linewidth]{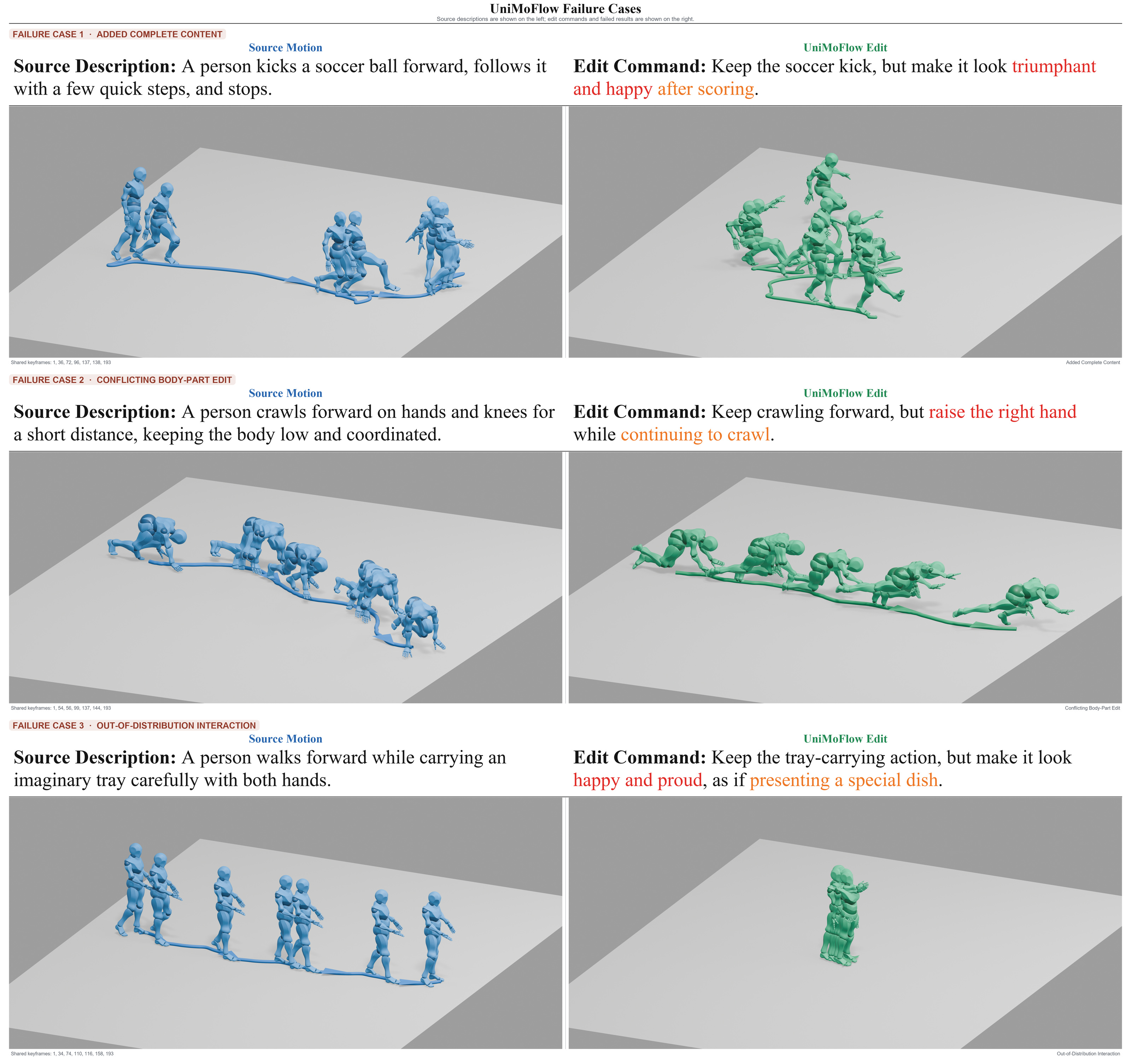}\caption{Representative failures: an abstract continuation after dense soccer motion (row 1), a right-hand edit conflicting with crawling support (row 2), and an affective object-interaction beyond the model's facial and semantic coverage (row 3).}\label{fig:failure-cases}\end{figure*}\section{Additional Qualitative Results}
\label{app:qualitative}
Following the main-paper protocol, Figure~\ref{fig:supplementary-comparison} applies identical Omni-MoEdit test sources and instructions to UniMoFlow and competing editors. The seven rows test both single and compound edits: reversing a turn while adding two right-arm waves; replacing crossed arms with an open, confident pose, hands on hips, and a leftward head turn; changing the torso-tilt direction before stepping left to recover; bending the knees during a sway; inserting a final left-foot backward step; replacing a forward lean with a backward lean while raising bent arms to the left; and inserting a forward step before a forward lean with raised arms. Within each instruction, red and orange phrases identify distinct semantic clauses, and dashed boxes of the same color mark the corresponding temporal and body regions in the outputs. The annotations therefore expose whether a method realizes each requested component while retaining the surrounding source motion; they are added only for visualization and are not supplied to the editors.

Figure~\ref{fig:five-basic-edits} isolates the five basic editing dimensions using one natural straight-forward walking sequence. The action-type branch changes walking into running with faster strides and stronger arm swings; the body-part branch preserves forward locomotion while raising and holding a right-hand salute; the amplitude branch enlarges both step length and arm-swing range; the timing branch preserves the initial walk and introduces a right turn followed by running only after several steps; and the style branch converts the gait into a crouched, cautious thief-like tiptoe. These controlled branches clarify the effect of each dimension, whereas realistic Omni-MoEdit commands frequently compose multiple dimensions, as illustrated in Figure~\ref{fig:supplementary-comparison}.

\subsection{Failure Cases}
The bottom panel of Figure~\ref{fig:failure-cases} summarizes three representative limitations. \textbf{Row 1: abstract continuation after a dense action.} The source already fills most frames with a dribble, run-up, shot, and stop; the abstract request to celebrate a goal provides neither a concrete celebratory motion nor an insertion point, producing an incoherent continuation. \textbf{Row 2: conflicting physical support.} Raising the right hand conflicts with the four-limb support required by crawling, so the lower body continues crawling while the hand floats implausibly. \textbf{Row 3: out-of-scope affective interaction.} Expressing a complex reaction to food while preserving an imaginary plate-holding action exceeds the facial and semantic coverage of the current representation and data, leading to an incorrect modification.

These cases show that current motion editing benefits from concrete kinematic instructions. Abstract requests become less reliable as source structure grows denser, and local edit semantics may conflict with global physical constraints. Temporal insertion, semantic abstraction, physical compatibility, and richer face--body--object interaction remain important directions for future work.

\paragraph{AI assistance disclosure.}
We used Codex with GPT-5.6 to assist with English-language polishing and to invoke Blender tools for motion-visualization rendering, skeletal rigging, and skinning. All research ideas, methods, experiments, analyses, and final manuscript decisions were produced and verified by the authors.


\end{document}